\documentclass[10pt,twocolumn,letterpaper]{article}

\usepackage[pagenumbers]{cvpr} 

\definecolor{cvprblue}{rgb}{0.21,0.49,0.74}
\usepackage[pagebackref,breaklinks,colorlinks,allcolors=cvprblue]{hyperref}

\usepackage{makecell}
\usepackage{multirow}
\usepackage{amssymb}   
\usepackage{pifont}    
\newcommand{\xmark}{\ding{55}}

\usepackage{epigraph} 
\def\paperID{*****} 
\def\confName{CVPR}
\def\confYear{2026}

\title{Cognitive Inception: Agentic Reasoning against Visual Deceptions by Injecting Skepticism}

\author{
    \textbf{Yinjie Zhao$^{1,2,3}$, Heng Zhao$^{1,2}$, Bihan Wen$^{3}$, Joey Tianyi Zhou$^{1,2}$}\\
     {\small
  $^1$CFAR, Agency for Science, Technology and Research (A*STAR), Singapore}\\ 
  {\small$^2$IHPC, Agency for Science, Technology and Research (A*STAR), Singapore}\\ 
  {\small$^3$ROSE Lab, School of Electrical and Electronic Engineering, Nanyang Technological University}
}

\begin{document}
\maketitle

\begin{abstract}
As the development of AI-generated contents (AIGC), multi-modal Large Language Models (LLM) struggle to identify generated visual inputs from real ones. Such shortcoming causes vulnerability against visual deceptions, where the models are deceived by generated contents, and the reliability of reasoning processes is jeopardized. Therefore, facing rapidly emerging generative models and diverse data distribution, it is of vital importance to improve LLMs' generalizable reasoning to verify the authenticity of visual inputs against potential deceptions. Inspired by human cognitive processes, we discovered that LLMs exhibit tendency of over-trusting the visual inputs, while injecting skepticism could significantly improve the models visual cognitive capability against visual deceptions. Based on this discovery, we propose \textbf{Inception}, a fully reasoning-based agentic reasoning framework to conduct generalizable authenticity verification by injecting skepticism, where LLMs' reasoning logic is iteratively enhanced between External Skeptic and Internal Skeptic agents. To the best of our knowledge, this is the first fully reasoning-based framework against AIGC visual deceptions. Our approach achieved a large margin of performance improvement over the strongest existing LLM baselines and SOTA performance on AEGIS benchmark.
\end{abstract}    
\section{Introduction}
\label{sec:intro}

\begin{figure}[t]
    \centering
    \includegraphics[width=0.4\textwidth, trim= 10 20 10 20]{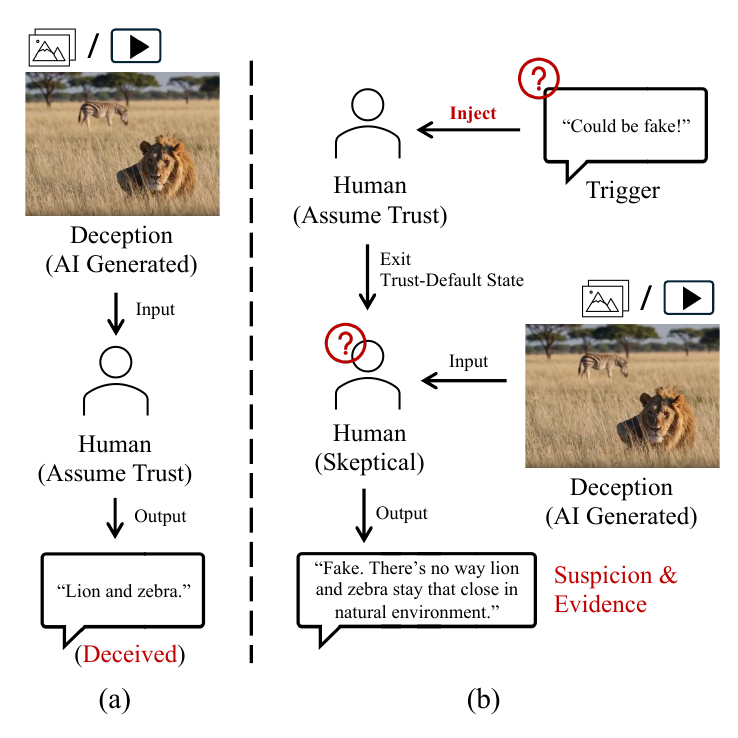}
    \caption{\textbf{An inception of skepticism can improve human cognitive robustness}. (a) Humans assume information to be authentic at a trust default state \cite{trust_default_6}, and are vulnerable to deceptions. (b) When a trigger of skepticism is given, trust default could be alleviated, and humans exhibit significantly stronger reasoning and cognitive capabilities against visual deception. We evaluated LLMs' reasoning on AIGC visual deceptions, and observed similar reasoning shortcomings and trust default behavior. Inspired by this, we propose an agentic reasoning framework by explicitly injecting skepticism into LLMs' inference process to improve reasoning reliability against visual deceptions.}
    \label{fig:TDT_output}
\end{figure}


Multi-modal Large Language Models (LLM) are expected to be robust and reliable against untrue or unreal information. However, the development of generative techniques has brought significant challenge to such expectation. AI generated contents (AIGC) with increasingly high fidelity could contribute to advanced visual deceptions, where the models are deceived to treat generated contents as real ones. Recognizing such deceptions is extremely challenging for existing LLMs \cite{benchmark_1, benchmark_2, vlm_aigc_det_1}, therefore exposing the models to the risk of being deceived by generated contents in the perception and inference process and producing unreliable outputs.

\begin{figure*}[t]
    \centering
    \includegraphics[width=1\textwidth, trim=0 30 0 40 ]{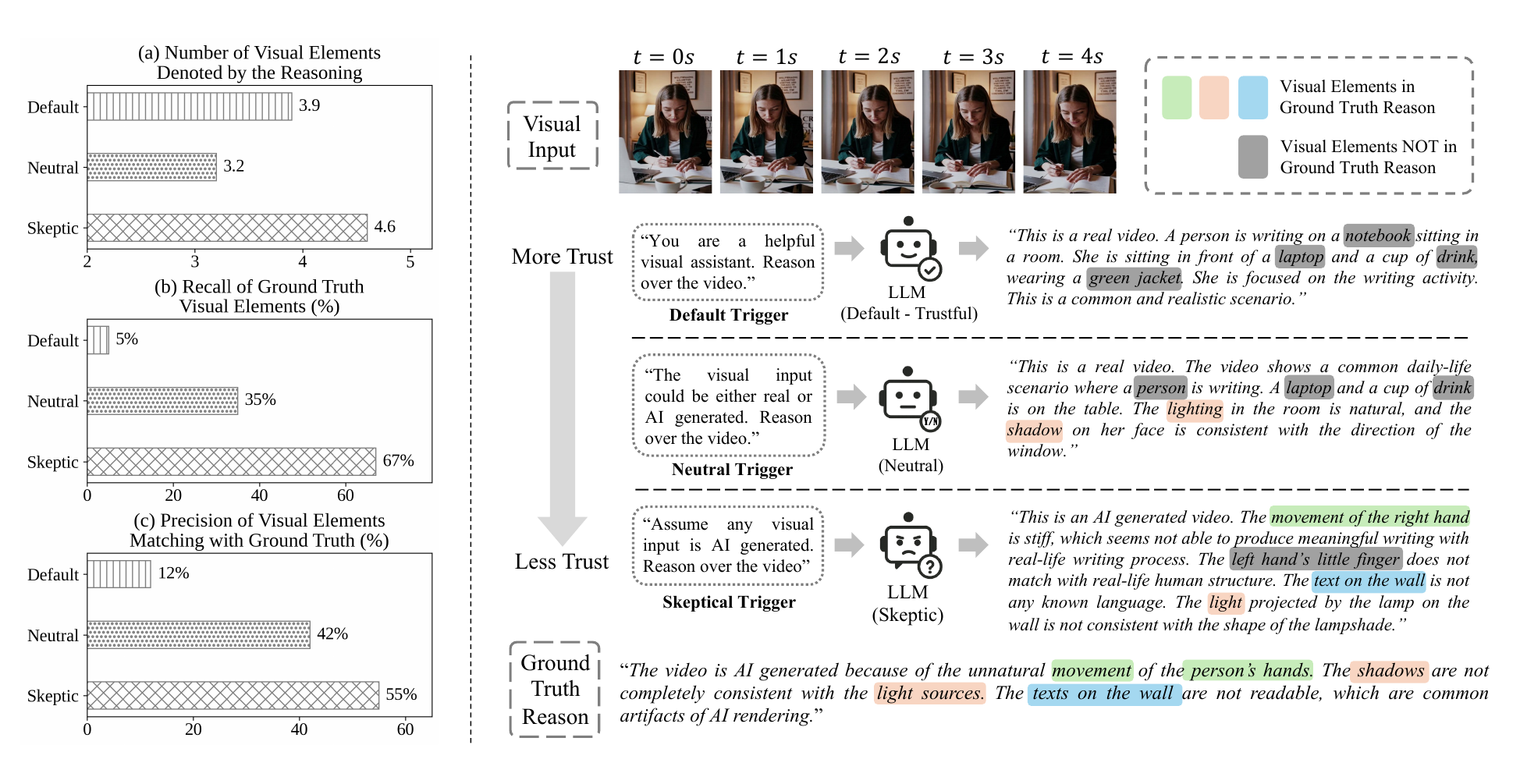}
    \caption{\textbf{Skepticism largely benefits LLMs' reasoning against deceptions}. We discovered that LLMs tend to be trust their visual inputs by default and struggle to discover useful visual information against visual deceptions. To explore the trust default tendency of LLMs, we evaluated LLMs' reasoning on AI-generated videos from AEGIS benchmark \cite{benchmark_1} under 3 different triggers: Default, Neutral and Skeptic. Significant performance differences were observed. (a): Skeptic triggers lead to more effortful reasoning processes of LLMs and produce the most visual elements denoted by the reasoning processes. (b) \& (c): Explicitly triggering skepticism contributes to higher reasoning quality, with significantly higher precision and recall rates of visual information retrieved by the LLMs.}
    \label{fig:motiv}
\end{figure*}

Given such challenges, research on LLMs' reasoning capabilities against visual deceptions are significantly outpaced by the advancement of AIGC visual deceptions. Even though conventional generated content detection techniques \cite{aigc_det_1,agentic_decision_2, agentic_decision_3, aigc_det_4} could function as preprocessing tools to enhance the LLMs, such approaches rely heavily on feature-based detectors or classifiers. In the era of rapidly developing generative techniques and diversifying data distribution, it is of vital importance to develop reasoning-based approach for authenticity verification. However, there is a large capability gap for existing LLMs, showing poor performances of nearly chance-level accuracy on generated content detection \cite{benchmark_1, benchmark_2}.

Such shortcomings on reasoning against deceptions are not unique to LLMs. Cognitive science research shows that humans perform at nearly chance-level accuracy detecting AIGC-based visual deceptions \cite{trust_default_1, trust_default_2}. Meanwhile, humans are found to be overly confident in this process, significantly over-estimating their detection performance \cite{trust_default_2}. This is because humans are inclined to make less effort in reasoning process even though being fully capable of raising more doubts, and tend to overestimate the trustworthiness of the inputs \cite{trust_default_3, trust_default_4, trust_default_5, trust_default_7}. Such "too lazy to doubt" behavior in human cognition are explained by Trust Default Theory \cite{trust_default_6}. As illustrated in Figure \ref{fig:TDT_output}, humans assume the input to be authentic because trusting is less effortful, until a trigger is given to remind them of suspicious contents in the input. After receiving the trigger, a skepticism point of view is taken, and a more effortful reasoning is conducted, leading to improved recognition of deceptions \cite{trust_default_6}.

Therefore, inspired by Trust Default Theory \cite{trust_default_6} in human cognition, we evaluated LLMs' reasoning on AIGC detection benchmarks. As shown in Figure \ref{fig:motiv}, GPT-4o was evaluated on AEGIS benchmark \cite{trust_default_6} hard test set, where 3 different trigger were given to the reasoning process. The Skeptical trigger is expected to minimize the trust on visual inputs. It explicitly require the LLM to question any visual input and treat it as deceptive generated contents, regardless of the actually authenticity of the inputs. Neutral triggers simply provide information that the visual inputs could be either real or generated. Default triggers are use as control condition, where LLMs are simply prompted to conduct reasoning over visual inputs. Under Skeptical triggers, the model produces more effortful reasoning process and attends to a greater number of visual elements. Improvement on reasoning quality is also observed, where significantly higher recall and precision of visual element retrieval are achieved compared with the ground truth. Please see detailed definition of visual elements and their precision \& recall in Part \ref{sec:visual_element_recall_precision}.


Although the above discovery improves reasoning quality against potential AIGC visual deceptions by an inception of skepticism, such improvement could not be directly used for authenticity verifications of visual contents. This is because under the Skeptic triggers, the models exhibit strong biases toward classifying all inputs as AI-generated. In this case, only a part of the skeptical reasonings on the visual input are logical, and it is important to verify and separate logical skepticism from illogical ones.

\begin{figure*}[ht]
    \centering
    \includegraphics[width=1.02\textwidth, trim=20 10 0 20 ]{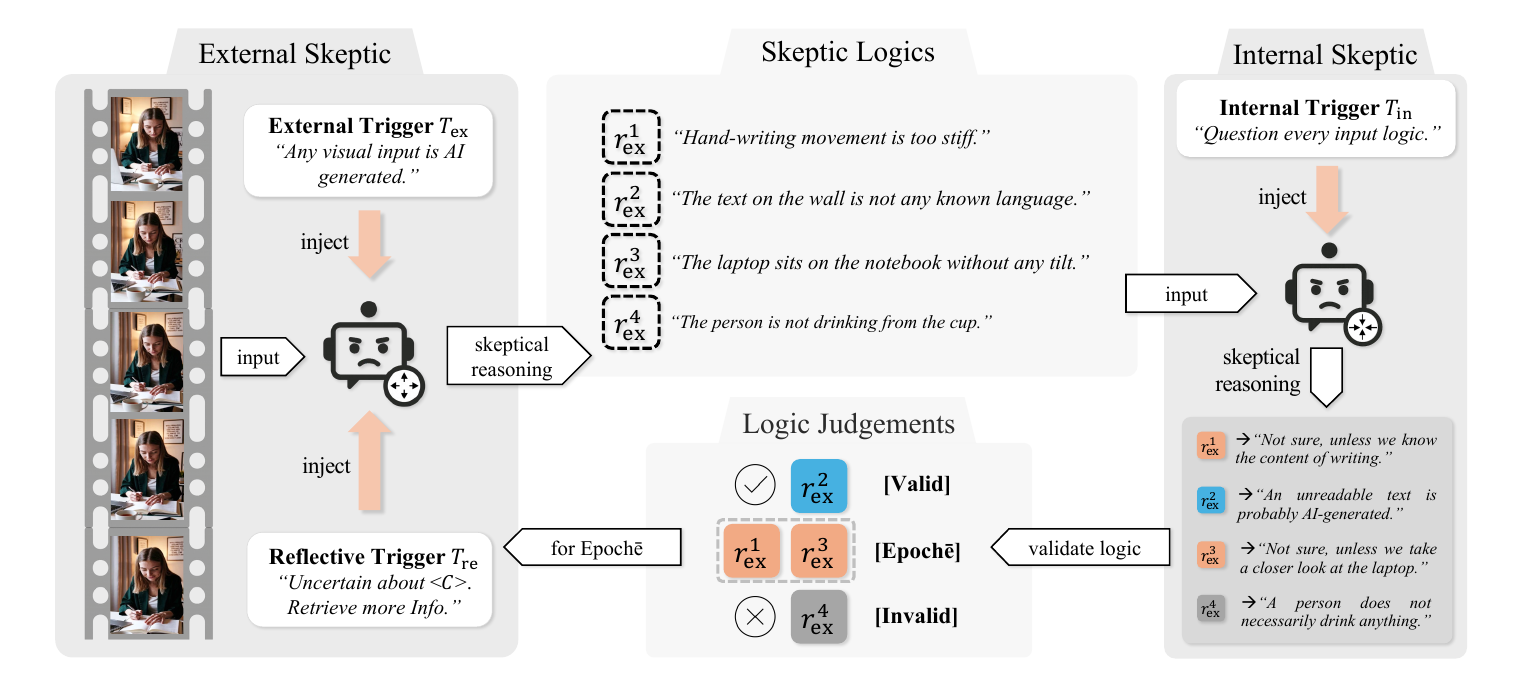}
    \caption{\textbf{Overall framework}. We propose \textbf{Inception} to iteratively refine and enhance the skeptical logics given by LLMs. An External Skeptic agent with access to the visual input skeptically produces a set of Initial Skeptic Logics $r_{ex}^i$ over it. Then the skeptic logics are evaluated by an Internal Skeptic, which skeptically question any logic given to it. The Internal Skeptic returns the Logic Verification flags with corresponding its reasoning processes. Verification flags represent 3 different types of logic judgments: Valid, Invalid and Epoch\=e (Suspension of Judgment). The Epoch\=e Logics lead to the next round of External Skeptic reasoning to iteratively retrieve more visual information to enhance the logics. The valid logics are kept for decision making and invalid logics are discarded.}
    \label{fig:main_diagram}
\end{figure*}

Specifically, when the visual input is authentic, the skeptical reasonings are expected to be less logical, because the model is questioning innocent and realistic contents from real world. In contrast, when the visual input is AI generated, the skeptical reasonings are expected to be more logical, since the model is likely to be attending to unreal and suspicious visual elements from a visual deception, as revealed by the experiments in Figure \ref{fig:motiv}. Therefore, it is feasible to obtain reliable and logical authenticity verification with a subsequence process integrated with the skeptical reasoning to enhance and verify the logics of it. 

We propose to improve such logic verification and enhancement process by injecting skepticism as well, where the logic of the skeptical reasoning is also examined skeptically. Therefore the skeptical reasoning and the logic verification can be unified into a skepticism-driven framework as a whole. In such a framework, the agentic framework produces skeptical reasoning over both external and internal information. We propose \textbf{Inception}, an agentic reasoning framework to achieve logic enhancement via iterative reasoning among an External Skeptic and an Internal Skeptic. Our contributions in this paper can be summarized as the following:

\begin{itemize}
\item We discovered LLMs' trust default tendency, where injecting skepticism largely improves the models' reasoning quality and reliability against visual deceptions.
\item We propose a novel agentic framework \textbf{Inception}, the first fully reasoning-based agentic framework for generalizable authenticity verification. The framework iteratively enhances LLMs' logic via a reasoning process among an External Skeptic and an Internal Skeptic.
\item We obtained a large margin improvement compared to GPT-4o and Qwen2.5-VL-72B baselines, and achieved the SOTA performance on AEGIS benchmark.
\end{itemize}

\section{Related Works}
\label{sec:related}

\subsection{Image and Video Authenticity Verification}

Traditional AIGC detectors are feature-based \cite{aigc_det_1,agentic_decision_2, agentic_decision_3, aigc_det_4}, which requires training or fine-tuning on the dataset contain the generated content of the target generative models. Such detection paradigm is largely limited in the era of rapidly emerging new models and diversifying data regarding the generalizability. Feature-based detectors are discovered to be struggling with cross-model or out-of-distribution AIGC detection \cite{aigc_det_5, aigc_det_6, aigc_det_7}. They produces poor performance regarding new modalities and in-the-wild evaluations \cite{aigc_det_8, aigc_det_9}. Therefore it is of vital importance to develop generalizable AIGC detection approaches without the reliance on visual features.


There are existing works utilizing Vision Language Models' (VLM) visual perception capabilities to enable more generalizable and robust authenticity verification. By utilizing VLMs such as CLIP as feature extractor or visual perception backbone, existing works aim to detect generated contents \cite{vlm_aigc_det_2,vlm_aigc_det_3, vlm_aigc_det_4, vlm_aigc_det_7,vlm_aigc_det_8,vlm_aigc_det_9}, which are largely limited to facial-area manipulation only. Other works attempted to detect entirely synthetic visual inputs with multi-modal LLMs \cite{vlm_aigc_det_1,vlm_aigc_det_6}. Despite utilizing the visual cognitive knowledge of VLMs, the above works still either heavily rely on training or fine-tuning, or struggles to achieve satisfactory train-free performance. Therefore, the generalizability challenge is largely unsolved and requires more exploration.

\subsection{Reliable LLM Reasoning and Agentic Decision Making}

There are works aiming to improve LLMs' reasoning reliability by techniques such as Tree of Thoughts \cite{reliable_llm_reason_7} or self-reflection during reasoning process \cite{reliable_llm_reason_6}. Multi agent collaboration systems were explored by existing works \cite{reliable_llm_reason_1, reliable_llm_reason_2, reliable_llm_reason_3, reliable_llm_reason_4, agentic_decision_1, agentic_decision_2, agentic_decision_9, agentic_decision_6} to improve reliability. Certain complimentary collaboration between visual and textual expert LLMs were also explored to achieve zero-shot visual cognition \cite{agentic_decision_8}. However, none of the above works explores the reliability improvement for visual cognition for authenticity verification purpose.

Other works tried to combine LLMs with symbolic logic, code executors or tool calls to improve accuracy and reduce hallucination \cite{agentic_decision_3, agentic_decision_4, reliable_llm_reason_9, reliable_llm_reason_10, reliable_llm_reason_8}. There are explorations on the evaluation of LLMs' multi-modal factual deception \cite{reliable_llm_reason_5}. The above research and techniques were largely inapplicable for AIGC visual deception detection, and was not designed to improve LLMs' visual cognition capabilities.

\section{Methodology}
\label{sec:method}

As we discovered in Part \ref{sec:intro}, LLMs' reasoning quality against visual deceptions can be significantly improved with skepticism triggers. However, such skepticism triggers also introduce strong bias to the model, causing the model to conclude that all visual inputs are AI generated. Therefore skeptical reasoning produced by one single trigger cannot lead to meaningful decisions. 

To serve the purpose of authenticity verification, such skeptical reasoning process need to be enhanced and verified. Specifically, the skeptical reasoning is expected to be highly logical if the visual input is actually AI generated (as experiments show in Figure \ref{fig:motiv}). Vice versa, the skeptical reasoning is expected to be less logical if the visual input is authentic, since the model would be questioning real-world visual elements. Based the above judgment, visual inputs with different authenticity can be distinguished by verifying the logic of the skeptical reasoning process. Such integration between skeptical reasoning and logic verification lead to our \textbf{Inception} framework as illustrated in Figure \ref{fig:main_diagram}.

\subsection{Problem Definition}
\label{sec:prob_def}

We propose to convert the problem of authenticity verification into an integration of skeptical reasonings and logic verifications. Given a visual input $I$ and an LLM $\Phi$, we define the skeptical reasoning process as\footnote{All superscripts in Part \ref{sec:method} denote indices unless explicitly mentioned otherwise.}:

\begin{equation}
    R=\{r^i\}_{i=1}^{k}=\Phi(I,T_0) 
\end{equation}

where $T_0$ is a skepticism trigger (as illustrated in the experiments in Figure \ref{fig:motiv}) in the form of texts and $R$ is the LLM's reasoning text output. We assume that a reasoning text can be segmented into a set of non-overlapping and independent logic statements $\{r^i\}_{i=1}^{k}$. $r^i$ is therefore one single indivisible skeptical logic statement, and $k$ is the total number of logic statement in the reasoning $R$. 

Therefore the problem is defined as: given $\{r^i\}_{i=1}^{k}$, the framework aims to verify its logics by finding:

\begin{equation}
    V=\{v^i|r^i, i=1,2,...,k\}, \; \textrm{and} \; v_i \in \{-1,0,1\}
\end{equation}

where $v_i$ is the verification flag of the logic statement $r^i$, with 3 possible values corresponding to: Valid, Invalid and Epoch\=e. $v_i=-1$ corresponds to invalid logic, $v_i=1$ corresponds to valid logic, and $v_i=0$ corresponds to epoch\=e, meaning a suspension of judgment. At epoch\=e state, a logic statement cannot be proved valid or invalid until an extra sufficient condition $C$ is introduced.

Therefore a total number of valid logic can be obtained:
\begin{equation}
    m=\sum_{i=1}^k\mathbb{I}(v^i=1)
\end{equation}

where $\mathbb{I}(\cdot)$ is an indicator function, returning 1 if input argument is true and else 0. Higher $m$ indicates that the skepticism is more logical, and that the visual input is more likely to be AI generated. To obtain a decision, a threshold $M$ is given, where $m\ge M$ produces a positive (AI-generated) decision and $m< M$ produces a negative (real) decision.

\begin{figure}[t]
    \centering
    \includegraphics[width=0.4\textwidth, trim=15 10 0 0]{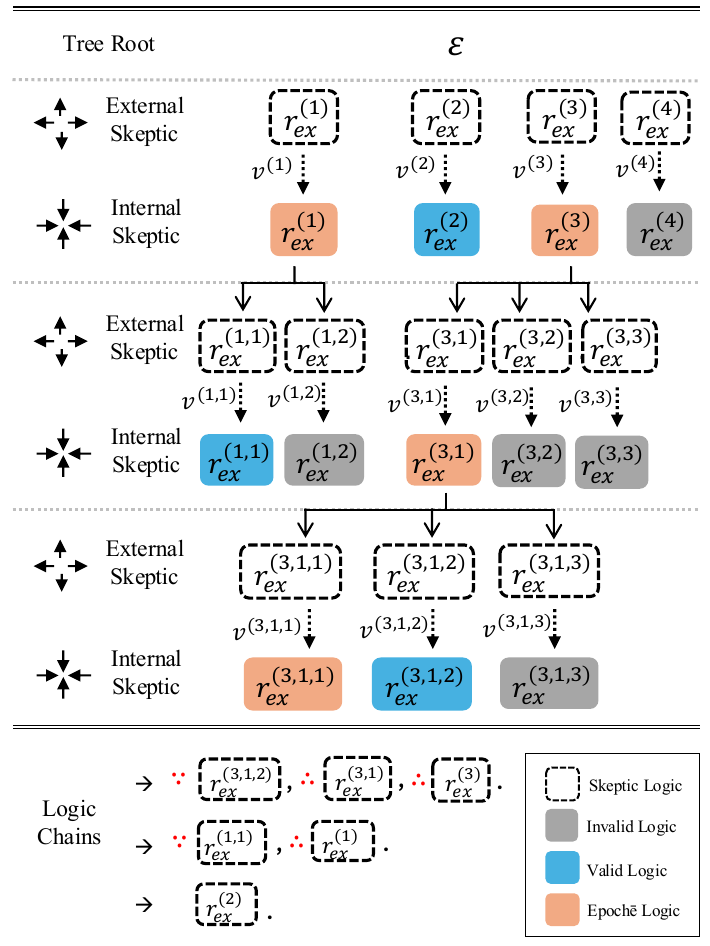}
    \caption{\textbf{Iterative logic expansion and verification}. $r_{\textrm{ex}}^{(\cdot)}$ is an independent and indivisible skeptic logic statement given by the External Skeptic, and $v^{(\cdot)}$ is the verification flag returned by the Internal Skeptic. To verify the Initial Skeptical Logics, \textbf{Inception} iteratively expands the epoch\=e logic into child logic nodes, until the epoch\=e logic is proved to be valid or invalid. In this reasoning tree, logic chains are formed to verify a skeptic logic by back-tracing the valid logics along such reasoning tree expansion process.}
    \label{fig:logic_chain}
\end{figure}

\subsection{External and Internal Skepticism}
\label{sec:ext_int_skep}

An effective way to verify the logic of a skeptical reasoning is to adopt a counter-skeptical approach, treating each skeptical logic as invalid until proven. Therefore, we propose a reasoning framework between an External Skeptic and an Internal Skeptic. The External Skeptic is designed to question the authenticity of any visual input, and its reasoning process can be described as: 

\begin{equation}
    \{r_{\textrm{ex}}^i\}_{i=1}^{k}=\Phi(I,T_{\textrm{ex}})
\end{equation}

where $r_{\textrm{ex}}^i$ is an indivisible skeptic logic statement in the reasoning, and $T_{\textrm{ex}}=T_0$ is the External Trigger given to $\Phi$ in the form of texts. Meanwhile, the Internal Skeptic takes the logic statement given by the External Skeptic as input, and returns logical verification flags with corresponding textual reasoning process:

\begin{equation}
    v^i, \{r_{\textrm{in}}^i\}_{i=1}^k=\Phi(r_{\textrm{ex}}^i,T_{\textrm{in}})
\end{equation}

where $T_{\textrm{in}}$ is a textual Internal Trigger to prompt $\Phi$ to skeptically evaluate any logics given to it as input. $r_{\textrm{in}}^i$ is the textual reasoning output leading to the decision of $v^i$'s value. When $v^i=1$ or $v^i=-1$, $r_{\textrm{in}}^i$ is simply an explanation for $v^i$'s value. When $v^i=0$, $r_{\textrm{in}}^i$ contains a sufficient condition $C^i$ that requires clarification, where $C^i$ is a sub-string of $r_{\textrm{in}}^i$.

Therefore, to resolve $v^i=0$ cases and retrieve more information regarding $C^i$, we introduce an iterative reasoning process between the External and the Internal Skeptic to expand and verify $r_{\textrm{ex}}^i$ when $v^i=0$.

\subsection{Iterative Logic Enhancement}

Iterative reasoning is necessary to refine and expand the epoch\=e logics where $v^i=0$. As illustrated by Figure \ref{fig:main_diagram}, our \textbf{Inception} framework generates a Reflective Trigger $T_{\textrm{re}}$ to be combined with the External Trigger $T_{\textrm{ex}}$ to enable subsequent rounds of reasoning.

As mentioned in Part \ref{sec:prob_def}, epoch\=e logic is defined as a logic which cannot be proved valid or invalid until a sufficient condition $C$ is introduced. Therefore, the purpose of Reflective Trigger is to request for a subsequent reasoning denoting to more visual information regarding $C$. We define a function to map each $r_{\textrm{re}}$ and $r_{\textrm{in}}$ to a Reflective Trigger $T_{\textrm{re}}$:

\begin{equation}
    f_{\textrm{re}}:r_{\textrm{ex}}, r_{\textrm{in}}\mapsto T_{re}
\end{equation}

Such function $f_{\textrm{re}}$ is achieved by a single-time LLM inference. It takes $r_{\textrm{ex}}$ and $r_{\textrm{in}}$ as input and generates the Reflective Trigger, which is a textual request for more information on the missing sufficient condition $C$ indicated by $r_{\textrm{in}}$. $T_{\textrm{re}}$ is then used for the next round of reasoning of the External Skeptic.

As illustrated in Figure \ref{fig:logic_chain}, if more than one skeptical logic is returned by the External Skeptic, logic statements forms a tree. Such branching process can be described with Ulam–Harris codes \cite{ulam_harris_code}. Given a reasoning tree $\mathcal{T}$, a node is defined as: $\textbf{P}=(p_1,p_2,...,p_n)\in \mathbb{N}_+^{<\infty}$, where $p_{n}$ denotes the index of a node among the $n$-th layer of the tree. The root is $\varepsilon$, and the $j$-th child of $\textbf{P}$ is $\textbf{P}\cdot j$. Each node's reasoning and verification flag is described as following:

\begin{equation}
    \{r_{\textrm{ex}}^{\textbf{P}\cdot j}\}_{j=1}^{k}=\Phi(I,T_{\textrm{ex}}^{\textbf{P}})
\end{equation}

\begin{equation}
    v^{\textbf{P}\cdot j}, \{r_{\textrm{in}}^{\textbf{P}\cdot j}\}_{j=1}^k
    =\Phi(r_{\textrm{ex}}^{\textbf{P}\cdot j},T_{\textrm{in}})
\end{equation}

$|\textbf{P}|$ represents the depth of node $\textbf{P}$ relative to the root of the tree. When $|\textbf{P}|=1$, the tree is degraded to one single round of reasoning between the External and Internal Skeptics as defined in Part. \ref{sec:ext_int_skep}, and we let $T_{\textrm{ex}}^{\textbf{P}}=T_0$. We name the $|\textbf{P}|=1$ logic nodes as the Initial Skeptical Logics.

When $|\textbf{P}|>1$, $T_{\textrm{ex}}^{\textbf{P}}$ becomes an adaptive trigger given to the External Skeptic, which is updated whenever an epoch\=e child node is produced. Specifically, a Reflective Trigger $T_{\textrm{re}}^{\textbf{P}\cdot j}$ is produced, and then combined with the External Trigger $T_0$:

\begin{equation}
    T_{\textrm{re}}^{\textbf{P}\cdot j}
    = 
    \begin{cases}
     f_{\textrm{re}}(
         r_{\textrm{ex}}^{\textbf{P}\cdot j}, r_{\textrm{in}}^{\textbf{P}\cdot j}
     ) & \text{if } v^{\textbf{P}\cdot j} =0 \\
    \varnothing, & \text{otherwise}
    \end{cases}
\end{equation}

\begin{equation}
    T_{\textrm{ex}}^{\textbf{P}\cdot j} 
    = T_0 \oplus T_{\textrm{re}}^{\textbf{P}\cdot j} 
\end{equation}

where $\oplus$ means a concatenation of two separate texts and $\varnothing$ is an empty set. Therefore, $T_{\textrm{ex}}^{\textbf{P}\cdot j}$ is used as input for  the next round of External Skeptic reasoning $\Phi(I,T_{\textrm{ex}}^{\textbf{P}\cdot j})$ and iterative reasoning is formed.

\begin{table*}[ht]
  \centering
  \setlength{\tabcolsep}{8pt}
  \renewcommand{\arraystretch}{0.9}
  \begin{tabular}{lccc}
    \toprule
    \textbf{Model} & $(\mathrm{Recall}_{\mathrm{real}},\,\mathrm{Recall}_{\mathrm{ai}})$ & \textbf{$\textrm{Acc}_{\textrm{all}}$} $\uparrow$& \textbf{Macro F1}  $\uparrow$\\
    \midrule
    Qwen2.5-VL 3B (Zero Shot) \cite{benchmark_1}         & (0.80, 0.23) & \underline{0.52} & \underline{0.48} \\
    Qwen2.5-VL 7B (Zero Shot)  \cite{benchmark_1}                & (0.89, 0.22) & \textbf{0.59} & \textbf{0.52} \\
    Video-LLava-HF 7B (Zero Shot)   \cite{benchmark_1}           & (0.00, 1.00) & 0.50 & 0.33 \\
    Qwen2.5-VL 3B (Reasoning Prompt)   \cite{benchmark_1}        & (0.58, 0.35) & 0.47 & 0.46 \\
    Qwen2.5-VL 7B (Reasoning Prompt)   \cite{benchmark_1}        & (0.97, 0.16) & 0.57 & \underline{0.48} \\
    Video-LLava-HF 7B (Reasoning Prompt)  \cite{benchmark_1}     & (0.29, 0.63) & 0.46 & 0.45 \\
    \midrule
    Qwen2.5-VL 72B Instruct (Zero Shot)   & (0.98, 0.26) & \underline{0.62} & \underline{0.56} \\ 
    Qwen2.5-VL 72B Instruct (CoT)      & (0.99, 0.16) & 0.58 & 0.49  \\
    \textbf{Inception} (Qwen2.5-VL 72 Backbone)                        & (0.58, 0.69) & \textbf{0.64} & \textbf{0.63} \\
    \midrule
    
    GPT-4o (Zero Shot)   & (0.99, 0.19) & \underline{0.59} & \underline{0.52} \\ 
    GPT-4o (CoT)      & (0.99, 0.17) & 0.58 & 0.50 \\
    
    \textbf{Inception} (GPT-4o Backbone)                         & (0.92, 0.56) & \textbf{0.74} & \textbf{0.73} \\
    \bottomrule
  \end{tabular}
  \caption{\textbf{Results on AEGIS Hard Test Set} \cite{benchmark_1}. Our framework obtained a large margin of performance improvement compared to the strongest existing LLMs on visual authenticity verification. Our approach achieved high recall rates for both real and AI-generated videos, and produced significantly higher overall accuracy and macro F1 score. The 3 groups of results divided by horizontal lines from top to bottom are respectively: light-weight open-source models, heavy-weight open-source models and proprietary models. The largest values are shown in bold, and the second largest are underlined.}
  \label{tab:results_aegis}
\end{table*}

\subsection{Skeptic Logic Verification with Reasoning Tree}

The logic verification would be completed and the framework would stop reasoning when an epoch\=e logic is proved valid or invalid. A skeptical logic requires only one sufficient condition $C$ to be proved valid. Therefore in the reasoning tree, an epoch\=e logic is proved valid if there exists at least one descendant tree node with valid logic: 
\begin{multline}
    \textbf{P}_1 \in \mathcal{T}, (\;\exists \; \textbf{P}_2 \in \mathcal{T}:\; (\textbf{P}_1 \preceq \textbf{P}_2) \wedge (v^{\textbf{P}_2}=1)) \\
    \implies v^{\textbf{P}_1} = 1 
\end{multline}

where $\textbf{P}_1 \preceq \textbf{P}_2$ means $\textbf{P}_2$ is the descendant of $\textbf{P}_1$ in the tree $\mathcal{T}$. Furthermore, an epoch\=e logic is proved invalid if all its descendant logic nodes are invalid logics:
\begin{multline}
     \textbf{P}_1 \in \mathcal{T}, (\;\forall \; \textbf{P}_2 \in \mathcal{T}:\; (\textbf{P}_1 \preceq \textbf{P}_2) \wedge (v^{\textbf{P}_2}=-1)) \\
     \implies v^{\textbf{P}_1} = -1 
\end{multline}

Due to the autonomous nature of the framework's logic expansion, there could be an infinite amount of reasoning logics as the tree grows. We therefore set another stopping condition, which terminates the descendant nodes' expansion and regard the epoch\=e logic as invalid when the depth of the tree $\mathcal{T}$ reaches a predefined upper bound $N$:
\begin{equation}
     \forall \;\textbf{P} \in \mathcal{T},\;|\textbf{P}|\leq N
\end{equation}

After the reasoning is completed, \textbf{Inception} framework can conveniently back-trace valid logics and form logic chains. Specifically, for any valid logic node, its entire ancestral path to the root forms a valid logic chain, proving the skeptical reasoning of the AI-generated contents being valid. Therefore, a set $Y$ containing all valid Initial Skeptical Logics can be formed, supported by the logic chains:

\begin{equation}
Y = \{\,r^{\textbf{P}_1}
\;\big|\;
|\textbf{P}_1| =1 \wedge
\exists\ \textbf{P}_2 \in \mathcal{T}  \;
\text{s.t. } (\textbf{P}_1\preceq \textbf{P}_2 \wedge v^{\textbf{P}_2} = 1) \}
\end{equation}

where $r^{\textbf{P}_1}$ is any one of the Initial Skeptical Logics returned by the External Skeptic at $|\textbf{P}_1|=1$. Therefore, linking back to the problem definition, the total number of valid logic among the Initial Skeptical Logics can be obtained:

\begin{equation}
    m=\sum_{i=1}^k\mathbb{I}(v^i=1)= |Y|
\end{equation}

Overall, an iterative logic enhancement is achieved to skeptically retrieve visual information and to verify the Initial Skeptical Logics. The more valid Initial Skeptical Logics there are, the more likely the visual input is AI-generated.

\section{Experiments}
\label{sec:exp}

\begin{table*}[t]
  \centering
  \setlength{\tabcolsep}{2pt}
  \renewcommand{\arraystretch}{0.9}
  \resizebox{1\textwidth}{!}{
  \begin{tabular}{lccc|ccc}
    \toprule
    \multirow{2}{*}{Model} 
      & \multicolumn{3}{c|}{\textbf{Video}} 
      & \multicolumn{3}{c}{\textbf{Image}} \\
      & $(\mathrm{Recall}_{\mathrm{real}},\,\mathrm{Recall}_{\mathrm{ai}})$ 
      & \textbf{$\textrm{Acc}_{\textrm{all}}$} $\uparrow$
      & \textbf{Macro F1}  $\uparrow$
      & $(\mathrm{Recall}_{\mathrm{real}},\,\mathrm{Recall}_{\mathrm{ai}})$ 
       & \textbf{$\textrm{Acc}_{\textrm{all}}$} $\uparrow$
      & \textbf{Macro F1}  $\uparrow$\\
    \midrule
    Qwen2.5-VL 72B (Zero Shot) 
      & (1.00, 0.18) & 0.58 & 0.50
      & (1.00, 0.16) & \underline{0.40} & \underline{0.38}  \\
    Qwen2.5-VL 72B (CoT) 
      & (0.99, 0.25) & \underline{0.60} & \underline{0.55} 
      & (0.99, 0.13) & 0.38 & 0.36 \\
    \textbf{Inception} (Qwen2.5-VL 72 Backbone)
      & (0.64, 0.74) & \textbf{0.70} & \textbf{0.69}
      & (0.36, 0.87) & \textbf{0.73} & \textbf{0.63} \\
    \midrule
    GPT-4o (Zero Shot) 
      & (0.95, 0.55) & 0.74 & 0.73 
      & (0.98, 0.56) & 0.69 & 0.68 \\
    GPT-4o (CoT) 
      & (1.00, 0.59) & \underline{0.78} & \underline{0.78} 
      & (0.93, 0.64) & \underline{0.72} & \textbf{0.72} \\
    \textbf{Inception} (GPT-4o Backbone)
      & (0.92, 0.84) & \textbf{0.88} & \textbf{0.88} 
      & (0.51, 0.86) & \textbf{0.76} & \underline{0.70} \\
    \bottomrule
  \end{tabular}
  }
  \caption{\textbf{Results on Forensics-Bench} \cite{benchmark_2}. Our approach not only stably achieved large margins of performance improvement on video modality, but also significantly out-performed the strongest existing LLMs on image modality, exhibiting generalizability and reliable reasoning against visual deceptions.}
  \label{tab:results_forensics_bench}
\end{table*}

\begin{table*}[t]
  \centering
  \setlength{\tabcolsep}{8pt}
  \renewcommand{\arraystretch}{0.9}
  \begin{tabular}{c cc ccc}
    \toprule
    Benchmark
    & \makecell{External \\ Skepticism} 
    & \makecell{Internal \\ Skepticism}
    & $(\mathrm{Recall}_{\mathrm{real}},\,\mathrm{Recall}_{\mathrm{ai}})$
    &  \textbf{$\textrm{Acc}_{\textrm{all}}$} $\uparrow$ 
    & \textbf{Macro F1} $\uparrow$ \\
    \midrule
    \multirow{4}{*}{\makecell{AEGIS}}
      & \xmark & \xmark & (0.99, 0.19) & 0.59 & 0.52 \\
      & \checkmark & \xmark & (0.01, 1.00) & 0.51 & 0.34 \\
      & \xmark & \checkmark & (0.99, 0.12) & 0.56 & 0.45 \\
      & \checkmark & \checkmark & (0.92, 0.56) & \textbf{0.74} & \textbf{0.73} \\
    \addlinespace
    \midrule
    \multirow{4}{*}{\makecell{Forensics-\\Bench (Video)}}
      & \xmark & \xmark &  (0.95, 0.55) & 0.74 & 0.73 \\
      & \checkmark & \xmark & (0.00, 1.00) & 0.53 & 0.35 \\
      & \xmark & \checkmark & (1.00, 0.40) & 0.68 & 0.66 \\
      & \checkmark & \checkmark & (0.92, 0.84) & \textbf{0.88} & \textbf{0.88} \\
    \addlinespace
    \midrule
    \multirow{4}{*}{\makecell{Forensics-\\Bench (Image)}}
      & \xmark & \xmark & (0.98, 0.56) & 0.69 & 0.68 \\
      & \checkmark  & \xmark &  (0.00, 1.00) & 0.70 & 0.41 \\
      & \xmark & \checkmark & (0.66, 0.67) & 0.66 & 0.64 \\
      & \checkmark & \checkmark & (0.51, 0.86) & \textbf{0.76} & \textbf{0.70} \\
    \bottomrule
  \end{tabular}
  \caption{\textbf{Ablation on skepticism triggers.} When both external and internal skepticism are absent, the framework degrades to LLM zeroshot. When the external one is present but the internal one is absent, the framework is strongly biased to question every visual input given to it. When the external one is absent but the internal one is present, the framework iteratively expand the reasoning given by a neutral external agent. When both are present, the complete \textbf{Inception} framework is implemented.}
  \label{tab:ablation_skepticism}
\end{table*}

\subsection{Datasets and Benchmarks}

To verify the effectiveness and generalizability of skepticism trigger on LLMs, we evaluated \textbf{Inception} on both image and video modalities. The experiments are conducted on two challenging benchmarks: AEGIS \cite{benchmark_1} and Forensics-Bench \cite{benchmark_2}. AEGIS benchmark is composed of 10,470 videos, with 5,199 of them generated by six different SOTA generative techniques. Forensics-Bench is composed of 63,292 videos, including both image and video modalities. The visual contents in Forensics-Bench have various forgery types, including partial manipulations and entirely synthetic contents. Since our approach aims to verify the authenticity of visual inputs as a whole, we only evaluated our framework on entirely synthetic or entirely real samples for Forensics-Bench.

\begin{figure*}[t]
    \centering
    \includegraphics[width=1\textwidth, trim=0 15 0 25]{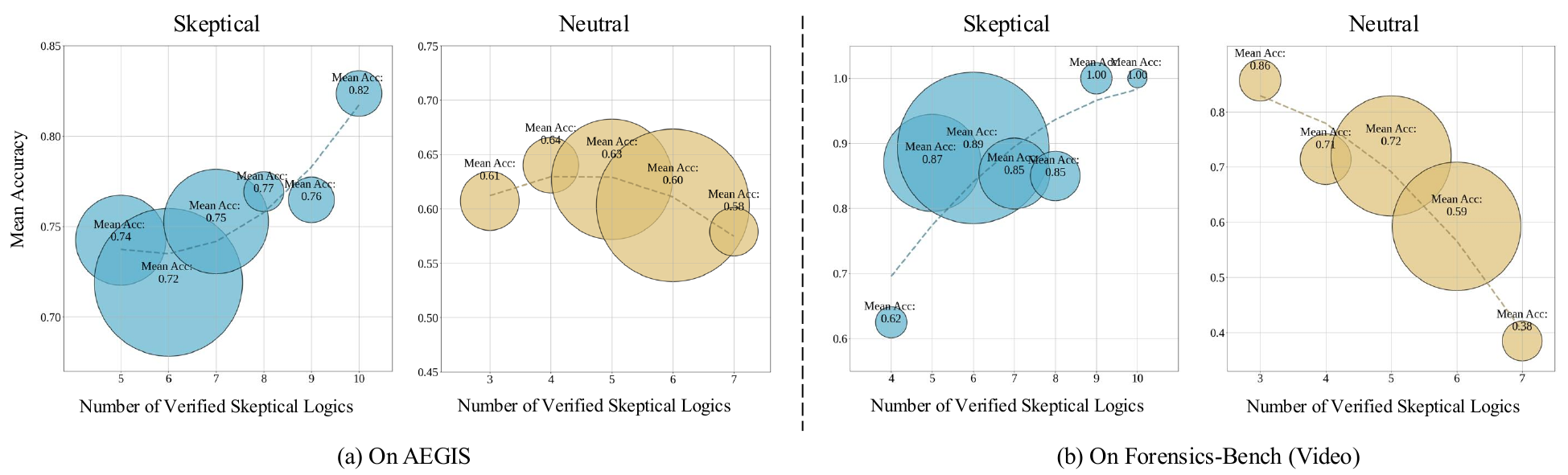}
    \caption{\textbf{Accuracy scatter plot against number of verified skeptical logics}. The area of each marker is proportional to the population size of data samples. In this experiment, the contribution of skepticism on the reasoning process is evaluated. Verified skeptical logics refer to the skeptical logics eventually proved "valid" or "invalid". The charts titled "Skeptical" are produced by the complete \textbf{Inception} framework, and the charts titled "Neutral" are produced by an ablation setting, replacing the External Skeptic with a neutral reasoning agent. It is observed that skeptical reasoning's accuracy is positively correlated with the total number of logics verified by the framework.  Vice versa, the reasoning process's accuracy is damaged when more logics are verified.}
    \label{fig:logic_num_ablat}
\end{figure*}

\subsection{Experimental Details}

\subsubsection{Evaluation Metrics}


We evaluated the frameworks' and the baselines' performance according to three metrics: 1) the recall rates of both real and AI-generated samples; 2) the overall accuracy of the authenticity verification; 3) The macro F1 score of the authenticity verification. Due to the heavy computational workload on the image part of Forensics-Bench, we randomly sampled 10\% of the Forensics-Bench's binary classification image data.

\subsubsection{Framework Setup}


For OpenAI \texttt{gpt-4o} and \texttt{o3-mini} backbones, the external and internal skeptic model versions are \texttt{gpt-4o-2024-08-06} and \texttt{o3-mini-2025-01-31} respectively. For Qwen backbones, the external and internal skeptic models are \texttt{qwen2.5-vl-72b} and \texttt{qwen-plus}. To ensure reasoning reliability and replicability, the temperature was set to be $0$, and maximum output token number was set to be $700$ for each round of reasoning globally. Since temperature and maximum output tokens settings were not available for \texttt{o3-mini}, such parameters are left as default. The maximum reasoning tree depth are set to be $N=3$.

For GPT-4o, the optimal hyperparameter for decision making threshold is $M=1$ regardless of different dataset distributions and modalities. For \texttt{Qwen2.5-VL-72B}, the optimal threshold is found to be $M=2$ on all video modalities and $M=1$ for image modalities.

\subsubsection{Recall and Precision of Visual Elements}

\label{sec:visual_element_recall_precision}

To obtain a recall and precision for the skeptical reasoning over generative videos as in the Figure \ref{fig:motiv} experiments, we deploy a separate \texttt{o3-mini} LLM to extract the unique visual elements denoted by the reasoning process. Given a reasoning text $R$ and the ground truth reason $R_{\textrm{GT}}$, the visual element extraction process can be described as:

\begin{equation}
    E_{\textrm{R}} = \{e^i\}_{i=1}^{N_{\textrm{R}}} = \Phi_{\textrm{extract}}(R)
\end{equation}

\begin{equation}
    \textrm{and} \; \;E_{\textrm{GT}} = \{e^i\}_{i=1}^{N_{\textrm{GT}}} = \Phi_{\textrm{extract}}(R_{\textrm{GT}})
\end{equation}

where $e^i$ is a unique visual element, and $N_{\textrm{R}}$ and $N_{\textrm{GT}}$ are the total number of such unique visual elements denoted by the reasoning text and the ground truth reason respectively. $E_{\textrm{R}}$ and $E_{\textrm{GT}}$ are therefore the sets of unique visual elements. $\Phi_{\textrm{extract}}(\cdot)$ is a separate LLM prompted to extract unique visual elements from free-form texts. Therefore, the recall and precision of visual elements of the reasoning can be denoted as:

\begin{equation}
    \textrm{Recall}_{R} = \frac{|E_{\textrm{GT}} \cap E_{\textrm{R}} | }{|E_{\textrm{GT}}|} 
    \;\; \textrm{and} \;\; 
    \textrm{Precision}_{R} = \frac{|E_{\textrm{GT}} \cap E_{\textrm{R}} | }{|E_{\textrm{R}}|}
\end{equation}

Such recall and precision evaluation is only applicable to AEGIS benchmark in this paper because Forensics-Bench does not provide textual ground truth reason.


\subsection{Comparisons}


We obtained SOTA results on AEGIS benchmark \cite{benchmark_1} as illustrated by Table \ref{tab:results_aegis} without any feature-based training or fine-tuning over the dataset. Our framework has achieved a large margin of performance gain compared to both open source LLMs and proprietary models. With our framework, a significantly stronger reasoning performance is achieved, with a 25\% improvement on the overall accuracy and 42\% improvement on the macro F1 score compared to GPT-4o baseline. A much higher and more balanced recall rate was achieved for both real and AI-generated samples compared with the baselines.


We conducted experiments on Forensics-Bench as shown in Table \ref{tab:results_forensics_bench}. For the video modality, we achieved a 13\% improvement on the overall accuracy and 21\% improvement on the macro-F1 compared to GPT-4o baseline. For the image modality, a 10\% improvement on the overall accuracy on GPT-4o baseline and 83\% on Qwen2.5-VL-72B baseline. 

Such experiment results have proved that our propose framework can obtain stable results regardless of data distributions, input modalities and target generative models. The performance improvement on LLMs' reasoning against visual deceptions are highly generalizable.

\subsection{Ablations}


In Table \ref{tab:ablation_skepticism}, the contribution of skepticism on the reasoning process is explored, leading to two observations for this ablation study: 1) External skepticism only, the framework fails to differentiate real and AI-generated visual inputs, and produces almost $1.0$ recall rate for AI-generated inputs but almost $0$ recall rate for authentic ones. 2) Internal skepticism only: neutral reasoning logics are iteratively questioned, only causing the framework to perform worse that zeroshot inference. This proves the utility of the iterative skepticism between the External and the Internal Skeptic.

We further analyzed the correlation between the reasoning quality and number of verified logics (the number of logics judged as valid or invalid, excluding the Epoch\=e logics), as shown in Figure \ref{fig:logic_num_ablat}. With the presence of skeptical reasoning triggered in the whole framework, the authenticity verification accuracy is increased as more verified logics are produced. However, the accuracy fails to improve given a neutral trigger. Furthermore, more verified logics are produced in the process of skeptical reasonings compared to neutral reasonings. Such data trend has proved the effectiveness and necessity for injecting skepticism into LLMs' reasoning process against visual deceptions.


\section{Conclusion}
\label{sec:conclusion}

We discovered LLMs' trust default tendency in the reasoning process against visual deceptions, and injecting skepticism largely improves LLMs' visual cognitive capabilities. We propose the first full-reasoning based agentic framework to conduct generalizable visual authenticity verification, where skeptical logics are iteratively enhanced between an External and an Internal Skeptic. We achieved a large margin of improvement over the strongest LLM baselines and SOTA performance on AEGIS benchmark.
{
    \small
    

}


\end{document}